\pdfoutput=1
\documentclass[11pt]{article}

\usepackage[final]{coling}

\usepackage{times}
\usepackage{latexsym}

\usepackage[T1]{fontenc}
\usepackage[utf8]{inputenc}

\usepackage{microtype}

\usepackage{inconsolata}

\usepackage{graphicx}

\title{MediaMind: Revolutionizing Media Monitoring using Agentification}

\author{Ahmet Gunduz \\
  \texttt{ahmet@aixplain.com} \\\And
  Kamer Ali Yuksel \\
  \texttt{kamer@aixplain.com} \\\And
  Hassan Sawaf \\
  \texttt{hassan@aixplain.com}\\}

\begin{document}
\maketitle
\begin{abstract}
In an era of rapid technological advancements, agentification of software tools has emerged as a critical innovation, enabling systems to function autonomously and adaptively. This paper introduces MediaMind as a case study to demonstrate the agentification process, highlighting how existing software can be transformed into intelligent agents capable of independent decision-making and dynamic interaction. Developed by aiXplain, MediaMind leverages agent-based architecture to autonomously monitor, analyze, and provide insights from multilingual media content in real time. The focus of this paper is on the technical methodologies and design principles behind agentifying MediaMind, showcasing how agentification enhances adaptability, efficiency, and responsiveness. Through detailed case studies and practical examples, we illustrate how the agentification of MediaMind empowers organizations to streamline workflows, optimize decision-making, and respond to evolving trends. This work underscores the broader potential of agentification to revolutionize software tools across various domains.
\end{abstract}

\section{Introduction}
Agentification refers to the process of equipping AI models, particularly large language models (LLMs), with autonomous capabilities, allowing them to operate as independent agents within a specified domain \citet{https://doi.org/10.48550/arxiv.2404.04442}. These agents are designed to generate and retain memories, enabling them to recall past interactions and decisions, which enriches their future responses and improves long-term task performance \citet{Wang_2024}. Additionally, they are endowed with access to specialized domain knowledge, enabling more accurate and context-specific outputs. The behavior and actions of an agent are guided through prompts that structure its interactions, ensuring that the LLM remains focused on the desired outcomes. The interface for interacting with these agents can be either a human agent or an AI agent, providing flexibility in how tasks are managed, with both human-like interaction and fully automated communication options. This allows agentified systems to operate seamlessly in environments requiring dynamic adaptability and precise knowledge, which in our case will be media monitoring.

\begin{figure}[t]
  \includegraphics[width=\columnwidth]{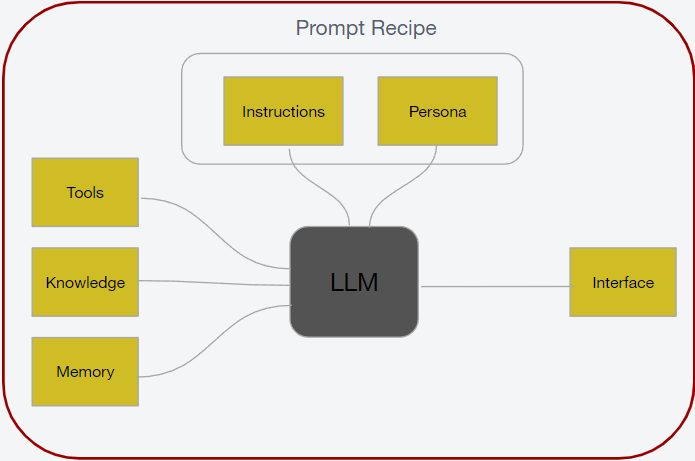}
  \caption{An LLM multi-agent system.}
  \label{fig:llm-agent}
\end{figure}

Media monitoring is the systematic process of continuously reading, watching, or listening to the editorial content of various media sources. It involves identifying, saving, and analyzing content that contains specific keywords or topics relevant to an organization’s interests. By capturing data across multiple platforms—including print, online, broadcast, and social media—media monitoring provides valuable insights into how a brand, industry, or specific issue is being discussed in the public domain.

These insights are critical for organizations, as media monitoring enables them to track their media presence, understand public sentiment, and anticipate market trends. With real-time awareness of public discourse, organizations can swiftly respond to potential crises, seize emerging opportunities, and maintain a competitive edge. Ultimately, effective media monitoring not only helps businesses protect their reputation but also shapes their public image and refines communication strategies, ensuring they stay ahead in a fast-evolving media landscape.

The landscape of media monitoring has become increasingly complex due to several key challenges. First, the overwhelming volume of media content generated every minute across various platforms presents a significant hurdle \citet{paine2011measure}. This data influx must be systematically tracked and analyzed, which traditionally required organizations to process vast datasets manually. Additionally, the diversity of content—ranging from multiple languages to different media formats such as text, video, and social media posts—compounds the complexity \citet{flew2014new}.

Beyond managing data volume, the challenge of accuracy looms large. Extracting meaningful insights from unstructured data, including slang, idiomatic expressions, and mixed languages, necessitates advanced tools for effective parsing \citet{macnamara2005media}. Moreover, the timeliness of responses is critical, as organizations must monitor media in real-time and receive alerts to address issues before they escalate \citet{gregory2010planning}. Equally important is depth; it’s not enough to track mentions—understanding sentiment, context, and thematic classification provides a deeper grasp of public discourse \citet{wright2010updated}.

These challenges highlight the need for transitioning from traditional tools to agentified systems, such as MediaMind by aiXplain. Agentification enables systems to autonomously process large datasets and make decisions on the fly, dramatically reducing costs and improving efficiency. For instance, instead of manually processing vast amounts of data, an agent can autonomously decide which sources or videos to analyze for sentiment, streamlining the process. Moreover, the user interface in an agentified tool becomes much more intuitive; rather than navigating dashboards and interpreting pre-defined charts, users can ask the agent for specific statistics, which it will retrieve and interpret directly from the database.

Through its agentic capabilities, MediaMind not only handles large volumes of diverse data but also enhances accuracy, timeliness, and depth in analysis, positioning it as a key solution in modern media monitoring. More importantly, its agentification marks a shift in how organizations can manage and leverage media insights effectively.

This paper provides a comprehensive exploration of MediaMind, an advanced media monitoring tool developed by aiXplain, designed to address the importance of transitioning into agentic systems. We will delve into the key features of MediaMind, highlighting how its AI-driven agentic capabilities enable organizations to effectively manage the vast volume and diversity of media content. The discussion will include an in-depth analysis of the AI pipelines that power MediaMind, demonstrating how these technologies facilitate accurate, real-time analysis of unstructured data across multiple languages and formats. Additionally, the paper will showcase the innovative aspects of MediaMind, such as its ability to provide deep insights into sentiment, context, and thematic classification, setting it apart from other media monitoring tools. Through this exploration, we aim to illustrate how MediaMind offers a robust solution to the complexities of tracking and analyzing media content in today’s dynamic landscape.

\section{Methodology}
\subsection{System Design and Workflow}
The backbone of the MediaMind system is built on the \citet{aixplain_repo}\footnote{\url{https://github.com/aixplain/aiXplain}} platform, which seamlessly integrates various AI models through its robust pipeline capabilities. For the proof of concept, YouTube was selected as the media source, providing a wide range of multimedia content for analysis. MediaMind features two versions: an older, traditional version where users manually carry out all steps of content analysis, and a new agentified version that automates these processes. 

The agentified MediaMind system comprises two critical pipelines: the **Video Analysis Pipeline**, which autonomously processes and analyzes media content, and the **Chat-Based Query and Answer Pipeline**, allowing users to interact with the system through natural language queries. This new version eliminates the need for manual intervention, enabling users to simply request insights or statistics, while the agent handles all the underlying tasks, enhancing efficiency and ease of use.

The backend of MediaMind hosts a variety of services, including AI analysis for processing and analyzing media content, data storage for persisting videos, analytics, and user settings, and user management for handling user accounts and permissions. Additionally, content scraping services are employed to download new content published daily. The API layer acts as a middleman, exposing endpoints such as "Search videos" for searching content, "Analyze content" for AI-driven content analysis, and "User settings" for managing user preferences and alerts. The frontend interacts with these endpoints by making HTTP requests, retrieving data, and rendering results on the user interface. This includes displaying processed content in the form of alerts, providing analytics dashboards, and delivering real-time notifications based on user-defined criteria.

\subsubsection{Video Analysis Pipeline}

The Video Analysis Pipeline exemplifies MediaMind's comprehensive analysis of video content, enabling users to derive insights from multimedia inputs. This pipeline consists of the following key stages:

\begin{itemize}
    \item \textbf{Extract Data:} The system extracts data from text, audio, image, and video inputs using advanced AI tools.
    \item \textbf{Speech Recognition:} Audio data from the video is transcribed into text using speech recognition models.
    \item \textbf{Sentiment Analysis:} The transcribed text is analyzed to determine sentiment and gauge public perception.
    \item \textbf{Named Entity Recognition (NER):} The text is processed using NER models to identify specific entities such as people, organizations, and locations.
    \item \textbf{Summarization:} Key points from the video are extracted and summarized, providing concise insights to the user.
\end{itemize}

\begin{figure}[t]
  \includegraphics[width=\columnwidth]{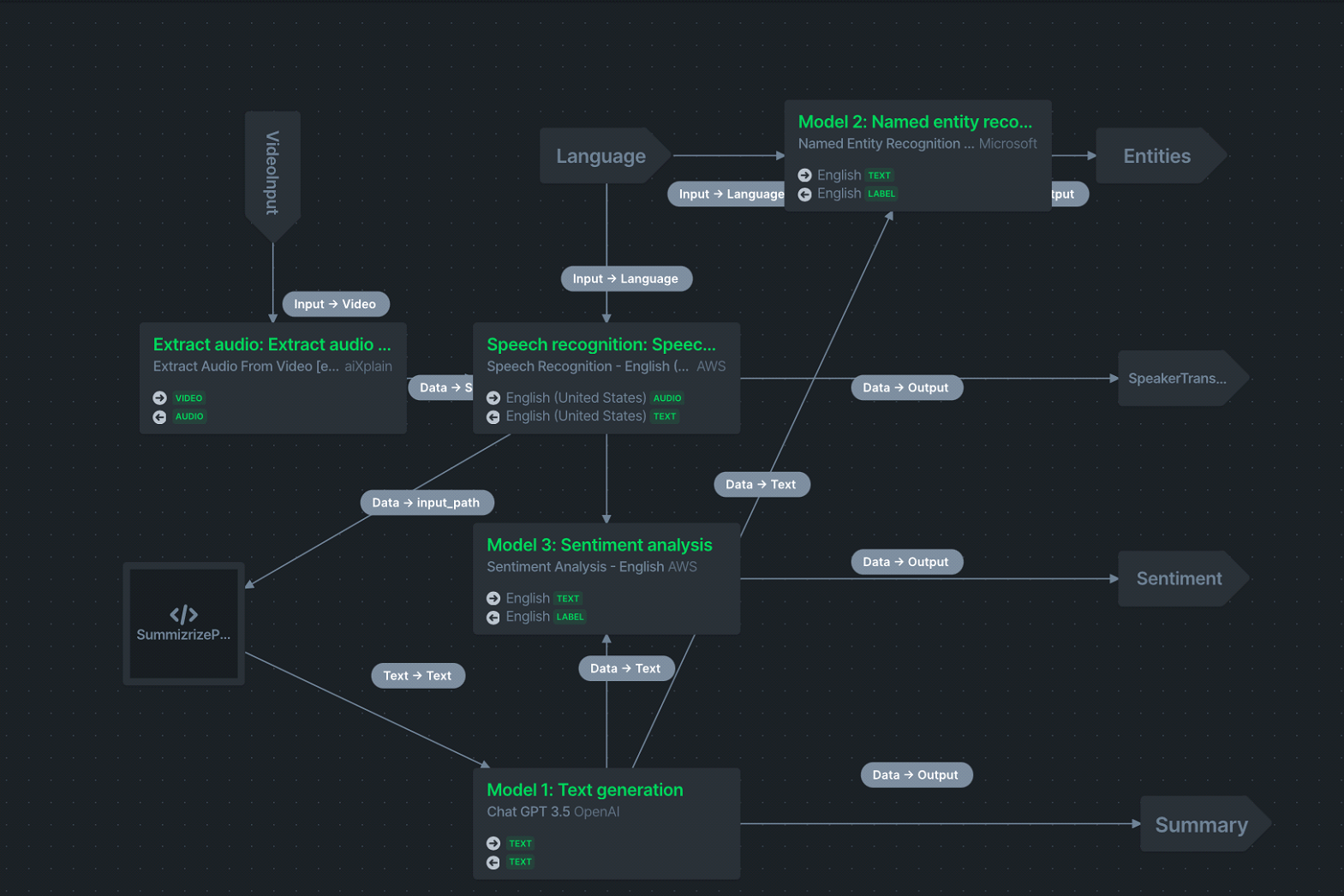}
  \caption{Video Analysis Pipeline Created in aiXplain.}
  \label{fig:aix-pipeline}
\end{figure}

Figure \ref{fig:aix-pipeline} illustrates the design of the video analysis pipeline developed using the Design feature in aiXplain platform.

\subsubsection{Chat-Based Query and Answer Pipeline}

The Chat-Based Query and Answer Pipeline highlights MediaMind's ability to respond to user queries with detailed, contextual answers. This pipeline operates in the following steps:

\begin{itemize}
    \item \textbf{Context Input:} User context and queries are input into the system, defining the scope of the analysis.
    \item \textbf{Query Generation:} AI models generate relevant queries based on the user's input and context.
    \item \textbf{Answer Generation:} Detailed answers are generated using the contextual understanding provided by AI models.
    \item \textbf{Output Processing:} The generated answers are processed and presented in a user-friendly format for easy interpretation.
\end{itemize}

\begin{figure}[t]
  \includegraphics[width=\columnwidth]{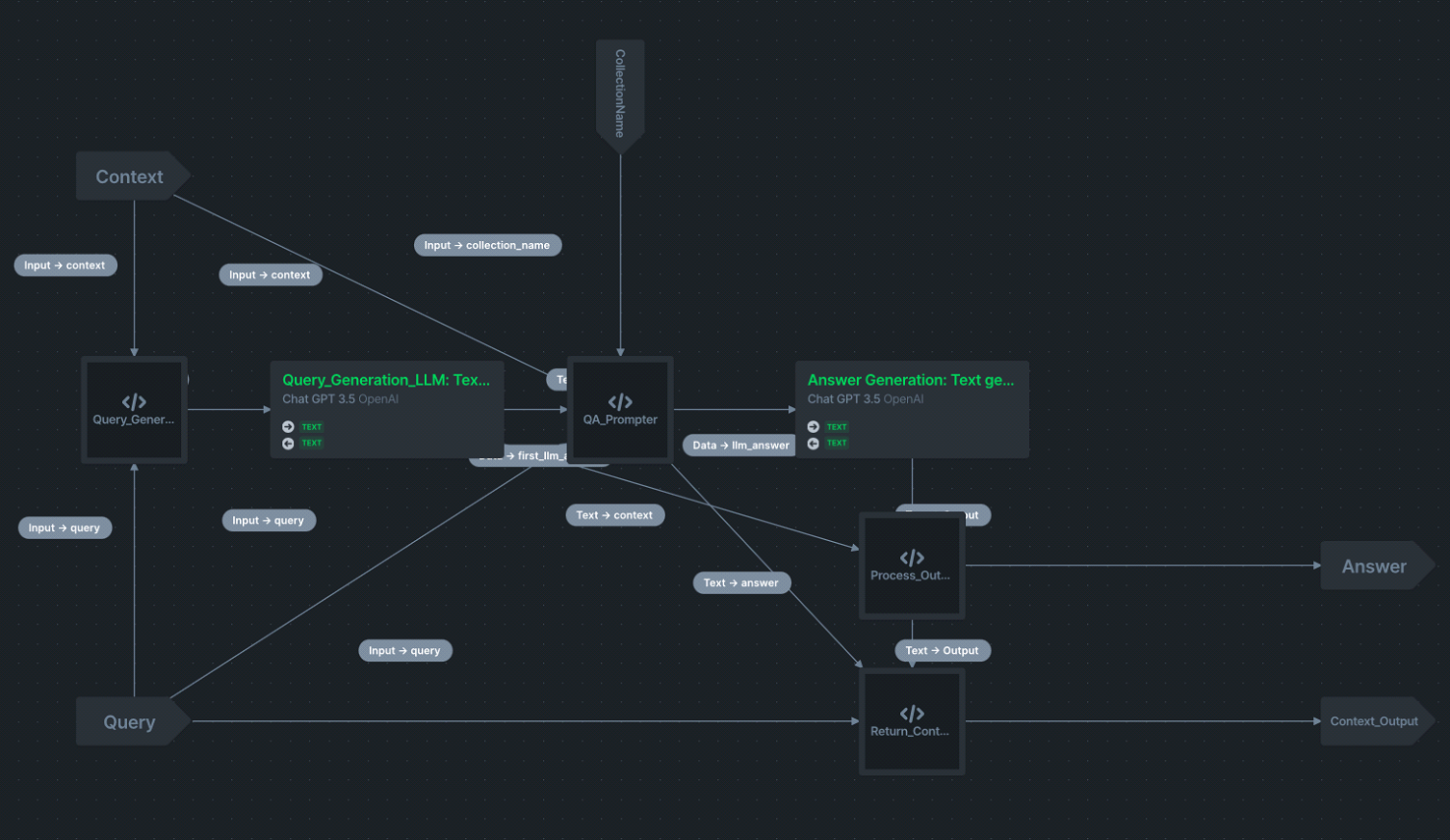}
  \caption{Chat-Based Query and Answer Pipeline Created in aiXplain.}
  \label{fig:aix-pipeline2}
\end{figure}

Figure \ref{fig:aix-pipeline2} depicts the design of the Chat-Based Query and Answer Pipeline, also constructed using the aiXplain platform.

\subsection{Agentic Capabilities}

In the first version of MediaMind, the application did not incorporate agentic capabilities. However, in the second version, significant enhancements were made by wrapping aiXplain pipelines and models to introduce them as tools for an agentic framework. This shift enabled more dynamic and flexible interactions within the system, allowing for deeper user engagement.

\begin{figure}[t]
  \includegraphics[width=\columnwidth]{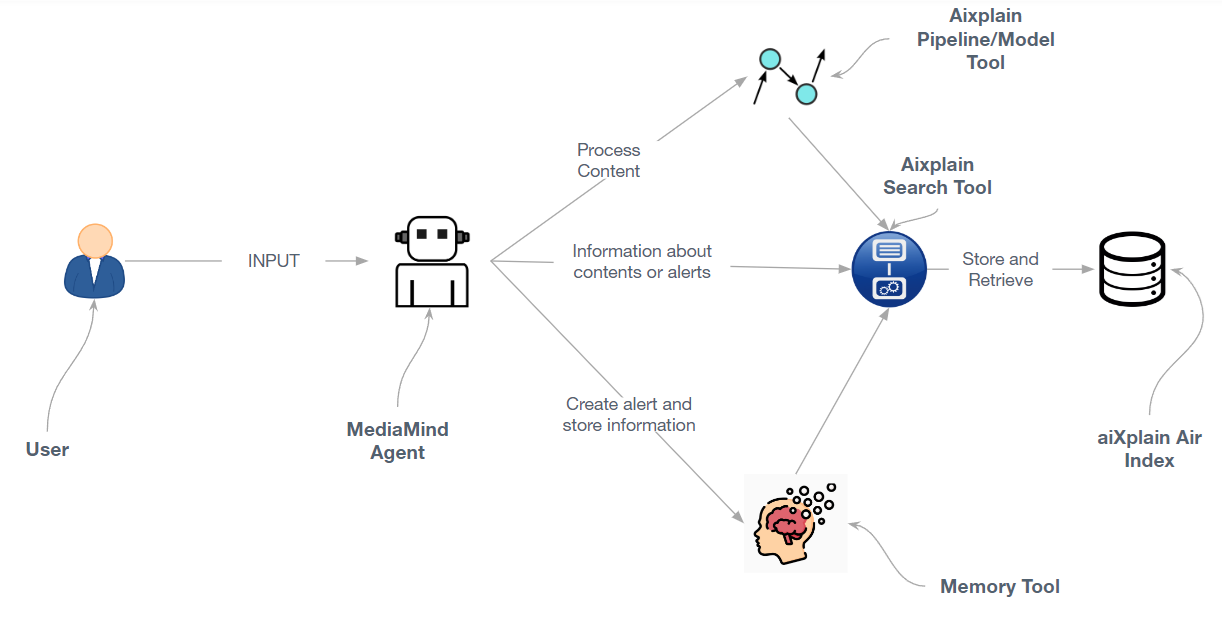}
  \caption{Agentifying MediaMind.}
  \label{fig:media-mind-agent}
\end{figure}

The MediaMind system incorporates agentic capabilities by utilizing a MediaMind Agent to autonomously process user input and interact with various tools from the aiXplain platform. The workflow, as illustrated in Figure \ref{fig:media-mind-agent}, follows these key steps:

\begin{enumerate}
    \item \textbf{User Input:} The workflow begins when a user provides input to the MediaMind Agent. This input can include queries about content, requests for media analysis, or commands to create alerts based on specific topics, sentiments, or other filters.

    \item \textbf{Processing by MediaMind Agent:} The MediaMind Agent serves as the core component, receiving the user’s input and determining the necessary operations. Based on the request, the agent interacts with the appropriate tools to retrieve or process media content.

    \item \textbf{aiXplain Pipeline/Model Tool:} For content analysis, the MediaMind Agent uses aiXplain’s Pipeline/Model tools to process the input content. This involves steps such as data extraction, speech recognition, sentiment analysis, and named entity recognition. The processed content can then be summarized and prepared for further analysis.

    \item \textbf{aiXplain Search Tool:} The MediaMind Agent also interacts with the aiXplain Search Tool, which allows it to search for relevant media content or alerts stored in the system. The Search Tool helps the agent retrieve stored data or retrieve new media relevant to the user’s input.

    \item \textbf{aiXplain Air Index:} The aiXplain Air Index serves as the data storage component. Once the agent has processed content or created alerts, the information is stored in the Air Index for future retrieval. This allows for efficient access to both new and previously processed media content.

    \item \textbf{Memory Tool:} The Memory Tool is used by the MediaMind Agent to store user-specific information, preferences, and past interactions. This allows the agent to provide personalized responses and adapt its behavior based on the user's history, ensuring a more intuitive and user-friendly experience.

    \item \textbf{Output to User:} After processing the content and retrieving the necessary information, the MediaMind Agent delivers the output to the user. This output could include processed media content, analytics, or new alerts based on the user’s initial input.
\end{enumerate}

By incorporating these aiXplain tools, the MediaMind Agent autonomously handles complex tasks such as content analysis, alert creation, and personalized responses, significantly streamlining the workflow. The agentic capabilities not only reduce manual intervention but also enhance the system’s responsiveness and flexibility in real-time media monitoring and analysis.

\section{MediaMind UI}

\begin{figure}[t]
  \includegraphics[width=\columnwidth]{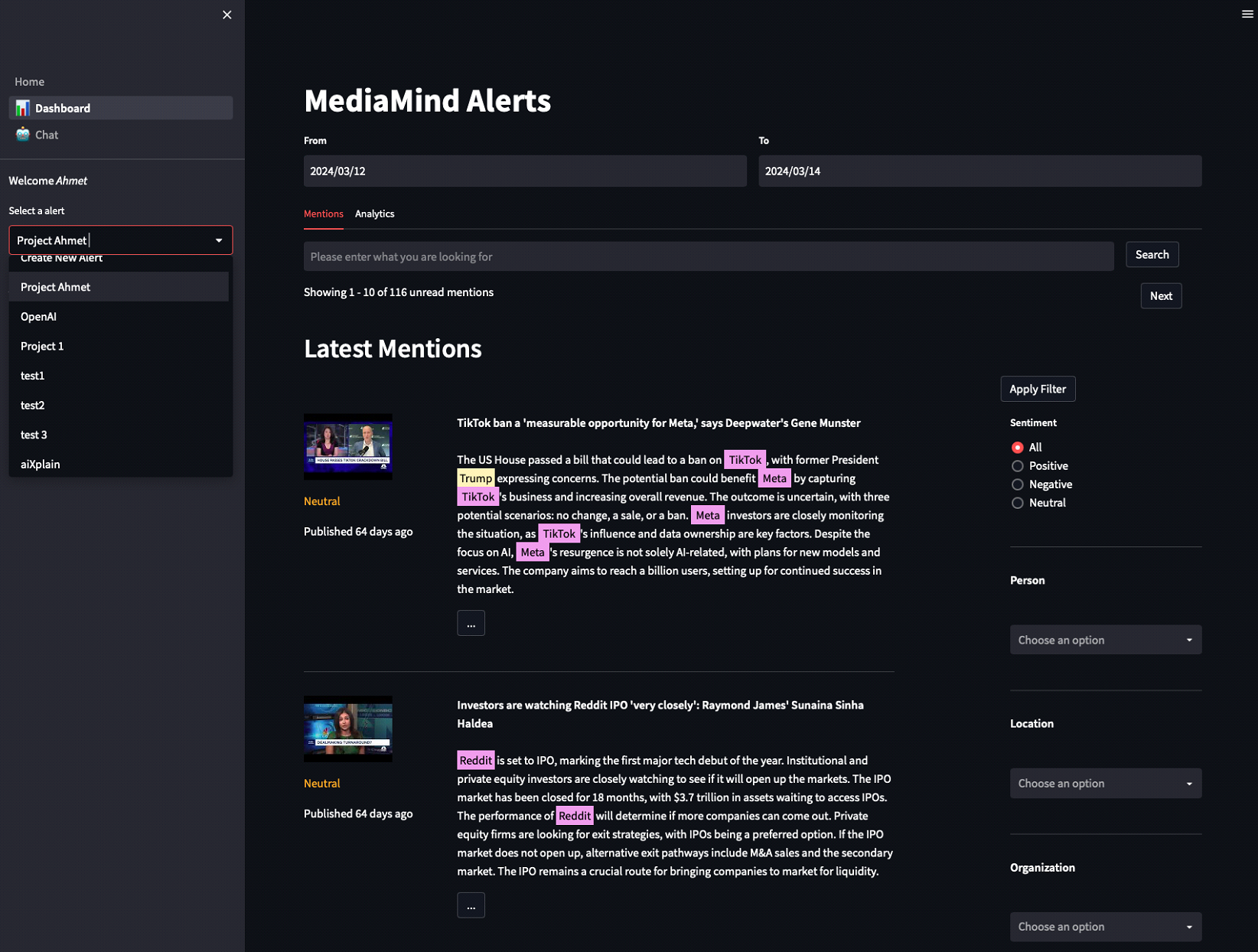}
  \caption{The User Interface of MediaMind.}
  \label{fig:media-mind-ui}
\end{figure}

The MediaMind user interface (UI), as shown in Figure \ref{fig:media-mind-ui} was developed using Streamlit\footnote{\url{https://github.com/streamlit/streamlit}}, providing a seamless and interactive experience for media monitoring. The UI consists of two primary pages: the \textbf{Dashboard} and the \textbf{Chat}.

\subsection{Dashboard}

The Dashboard page allows users to manually create alerts by specifying key details for monitoring media content. Users can define the \textbf{topics} they wish to monitor, select the \textbf{time period} for the media to be tracked, and apply various filters such as \textbf{sentiment} (positive, negative, or neutral), as well as more granular options like \textbf{location}, \textbf{person}, or \textbf{organization}. 

The Dashboard is divided into two main sections:
\begin{itemize}
    \item \textbf{Mentions:} This section displays the latest mentions of the specified topics, with links to relevant content. In the case of this implementation, these mentions are primarily links to YouTube videos.
    \item \textbf{Analytics:} This section provides various analytical insights related to the created alert. The analytics include the total \textbf{number of mentions}, breakdown of \textbf{positive} and \textbf{negative sentiments}, and visualizations such as \textbf{number of contents} over time.
\end{itemize}

\subsection{Chat}

The Chat page introduces agentic capabilities to MediaMind. This feature allows the user to interact with the system through a chat interface, powered by a large language model (LLM). By simply engaging in conversation, the user can autonomously perform the tasks that would normally be done manually in the Dashboard. For example, the user can request the creation of alerts, specify filters, and ask for analytics, all through natural language queries.

This is achieved by integrating aiXplain pipelines, which were wrapped and introduced as tools for the LLM agent. Through this agentification, the LLM is able to use the aiXplain pipelines to extract data, analyze sentiment, and present the results in a dynamic and interactive manner, streamlining the media monitoring process.

\section{Discussion}

The development of MediaMind represents a significant advancement in the realm of media monitoring, offering innovative technical capabilities that address many of the challenges associated with tracking and analyzing media content. Through a combination of multilingual support, multimodal content analysis, and advanced AI-driven analytics, MediaMind provides users with comprehensive insights into their media presence across a variety of formats and languages.

\subsection{Technical Innovations Behind MediaMind}

One of the standout features of MediaMind is its ability to perform \textbf{global multilingual media monitoring}. By supporting over 40 languages, the platform ensures comprehensive coverage of media content, enabling users to track mentions across different regions and languages with ease. This is particularly beneficial for global organizations, public figures, or entities that require detailed analysis of media content from diverse sources.

MediaMind also excels in \textbf{advanced analytics}, employing AI-driven techniques such as entity recognition, sentiment analysis, topic clustering, and summarization. These capabilities offer users the ability to delve deep into the context and sentiment behind specific mentions, providing actionable insights that can inform strategic decisions and enhance media management.

Moreover, MediaMind's ability to conduct \textbf{multimodal content analysis} ensures that no format is overlooked. The platform’s ability to analyze text, audio, images, and video content provides a holistic view of media presence, capturing the full spectrum of media mentions and coverage.

The \textbf{content search service} powered by semantic search enhances the user’s ability to find relevant information within vast media libraries. This feature, combined with \textbf{Named Entity Recognition (NER)} and \textbf{Sentiment Analysis}, enables users to monitor the presence and perception of specific individuals or topics across various media formats, providing a comprehensive view of their media footprint.

In addition to these technical innovations, MediaMind offers \textbf{summarization} capabilities that allow users to generate concise summaries of video content. This feature enables quick comprehension of key points, saving time while maintaining productivity. 

Finally, the \textbf{intuitive user interface} and \textbf{custom alerts and reporting} features make MediaMind accessible and user-friendly. Real-time dashboards and custom alerts keep users updated on emerging mentions and trends, while customizable reports ensure that stakeholders receive relevant and comprehensive insights.

\subsection{Limitations}

Despite these significant innovations, MediaMind currently has limitations that need to be addressed. As a proof of concept, YouTube was the only media source integrated into the system. While YouTube offers a rich repository of media content, expanding MediaMind’s monitoring capabilities to other media sources, such as social media platforms, news websites, and other digital channels, would greatly enhance its coverage and utility.

\subsection{Future Work}

Looking ahead, several key areas are identified for future development to further enhance MediaMind’s capabilities. A primary focus will be on \textbf{expanding the range of media sources} monitored. Incorporating additional social media platforms, news websites, and other digital channels will allow for broader, more comprehensive media monitoring, ensuring that no mention is missed.

Another area of development will be in enhancing \textbf{automated reporting}. The goal is to provide users with more detailed, customizable reports that can be tailored to specific stakeholders' needs, enabling deeper insights and more effective media management.

Additionally, improving \textbf{real-time analysis and reporting} will be critical for users who need immediate updates and insights. Enhancements to the real-time analytics features will allow users to respond more quickly to media mentions and trends, offering faster reaction times and more proactive media management.

In summary, while MediaMind currently offers a powerful media monitoring solution with significant technical innovations, there remains ample room for growth and improvement. By expanding media source coverage, automating and enhancing reporting, and providing real-time analysis, future iterations of MediaMind will provide even more robust and comprehensive media monitoring capabilities.

\section{Conclusion}
This paper introduced MediaMind, an agentic media monitoring tool built on the aiXplain platform, designed to address the growing demands of multilingual, multimodal media analysis. By integrating advanced AI models for sentiment analysis, entity recognition, and summarization, MediaMind offers a comprehensive solution for tracking media coverage across multiple formats and languages. MediaMind's intuitive interface, combined with its ability to create custom alerts and provide detailed analytics, empowers users to efficiently monitor their media presence. While the current implementation is limited to YouTube as a media source, future developments aim to expand coverage to more platforms and enhance real-time reporting features. In conclusion, MediaMind represents a robust and scalable media monitoring solution, capable of evolving to meet the dynamic needs of modern media landscapes. Its agentic capabilities make it a forward-thinking tool, poised to streamline media tracking and analysis across global markets.

% Bibliography entries for the entire Anthology, followed by custom entries
%\bibliography{anthology,custom}
% Custom bibliography entries only
\bibliography{custom}

\end{document}